\documentclass[10pt,conference,letterpaper]{IEEEtran}

\usepackage[hyphens]{url}
\usepackage{cite}
\usepackage{graphicx}
\usepackage{booktabs}
\usepackage{amsmath}
\usepackage{amssymb}
\usepackage{array}
\usepackage{balance}
\usepackage{afterpage}

\newcolumntype{L}[1]{>{\raggedright\arraybackslash}p{#1}}
\newcounter{algorithm}

\title{What Does an LLM-Agent Leaderboard Rank Actually Compare?}

\author{
\IEEEauthorblockN{Wei-Jung Huang}
\IEEEauthorblockA{\textit{Independent Researcher}\\
United States\\
william.wj.huang@gmail.com}
}

\begin{document}
\maketitle

\begin{abstract}
An LLM-agent leaderboard invites a familiar inference: an agent ranked above another is the better agent. Public evaluation logs may not support that conclusion when systems differ in task mixture, label source, release detail, or cost rule. We study what leaderboard scores estimate and when they justify pairwise superiority conclusions. Our estimand-aware pairwise procedure states the comparison target and measurement source, checks common support, and evaluates the supported difference using a stated uncertainty rule and practical margin. Controlled checks evaluate the decision labels under known finite-sample conditions and show why uncertainty must be included when judging sensitivity to target reweighting. Across SWE-bench, AgentRewardBench, and tau2-bench, close rank differences are often unresolved; proxy labels and utility rules can also change which system is selected. DataAgentBench and Open Agent show what remains estimable from coarser public records. A leaderboard score summarizes a released evaluation, whereas a fine-grained superiority claim additionally depends on the estimand and uncertainty rule used to interpret the difference.
\end{abstract}

\section{Introduction}

Leaderboard ranks are often used as shortcuts for model choice: if row A is above row B, A is treated as the better agent. For LLM agents, that shortcut is fragile because a row summarizes not only a base model but a full configuration of scaffold, tools, action space, prompt policy, stopping rule, label source, and resource budget. A scaffold comprises the prompts, tools, and control logic around the base model. A label source is the mechanism that assigns success, such as tests, expert annotations, functional evaluators, or LLM judges. Recent benchmarks differ in domain, environment, horizon, label source, and scaffold \cite{agentbench,swebench,agentrewardbench,hal,dataagentbench,tau2bench}, and automatic judges add measurement assumptions of their own \cite{mtbench,fairllmjudge,agentrewardbench}. A displayed ordering therefore reflects a particular evaluation design, even when it is read as a general ordering of agents.

The problem is visible in public agent-evaluation releases. On AgentRewardBench, expert labels select Claude 3.7 Sonnet as the top agent, while labels from a GPT-4o-mini judge select GPT-4o \cite{agentrewardbench,agentrewardbenchdata}. On SWE-bench Verified, among the 45 pairs formed conditional on the observed raw top 10, descriptive pointwise 95 percent intervals and a 2-point decision rule give 5 stable, 1 target-sensitive, and 39 underpowered comparisons; a max-deviation simultaneous interval resolves none of the 45 \cite{swebench,swebenchexperiments}. On tau2-bench, pass@4, meaning success in at least one of four attempts, raises scores by 17.7 to 20.4 points relative to pass@1 among comparable primary runs, while cost changes which system a utility rule selects \cite{tau2bench,tau2benchdata}. Together, these examples show that a released score answers one evaluation question but may not support a broader ordering.

Adding more raw metrics helps only when each metric's estimand is stated. The unadjusted success rate is a valid score for the observed benchmark mix. It answers a different question from equal-repository weighting, expert-labeled success, deployment-like success rules, or cost-aware utility. The central problem is that published rankings often invite broader pairwise conclusions than their score estimands justify.

\begin{figure*}[t]
\centering
\includegraphics[width=\textwidth]{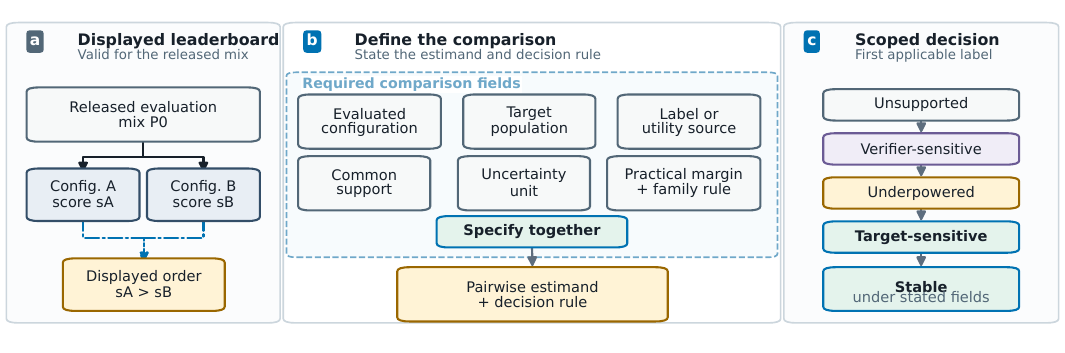}
\caption{From a displayed ordering to a scoped pairwise conclusion. Interpreting a raw score as pairwise superiority requires an evaluated configuration, target population, label or utility source, common support, uncertainty unit, practical margin, and comparison family when applicable. The procedure retains all diagnostic flags and reports the first applicable ordered label.}
\label{fig:estimand-overview}
\end{figure*}

Figure~\ref{fig:estimand-overview} summarizes the distinction between a released score and the additional conditions required for a pairwise superiority conclusion. A stable label applies only to the stated comparison fields; it does not identify causal effects of models, scaffolds, tasks, modes, or verifiers.

We introduce an estimand-aware pairwise decision procedure for public leaderboard ranks. An estimand is the quantity a score is intended to estimate, such as success under a stated task population, label source, or resource rule \cite{cochran1977sampling,little1993poststratification,imbensrubin2015}. The procedure treats a displayed rank as a pairwise superiority claim and asks whether the released records identify that claim. For any pair, it records the evaluated unit and target, checks common support, and assesses the supported gap using a stated uncertainty rule and practical margin. It reports an inferentially supported winner only when the evidence permits one; otherwise, it identifies the limiting condition: missing support, disagreement between expert and proxy labels, insufficient precision, or sensitivity to a target change within common support.

Before turning to public leaderboards, where the true ordering is unknown, we examine how the procedure behaves in controlled settings. One simulation tests whether the decision labels recover known states, another tests estimation under a known target, and an AgentRewardBench permutation analysis supplies a task-conditioned exchangeability reference. We then apply the same rule to the detailed records released for SWE-bench, AgentRewardBench, and tau2-bench. DataAgentBench and Open Agent expose coarser records, showing what score reweighting and grouped comparisons remain possible without item-level uncertainty.

Across these releases, target reweighting and uncertainty leave close comparisons unresolved more often than they reverse the order. We use \emph{weak verifier} for an automatic or functional label source used in place of expert or official success labels. Weak verifiers, repeated-attempt rules, costs, and aggregation can also change the selected system or the quantity being measured. These findings motivate reporting an estimand and a decision rule with each fine-grained ordering.

Prior work explains parts of this problem: benchmark composition, perturbations, and uncertainty can change conclusions \cite{benchmarktargets,siskabenchrobust,llmuq}; evaluation documentation makes reporting conditions explicit \cite{evalfactsheets}; and statistical comparison work supplies target-population and uncertainty tools \cite{cochran1977sampling,little1993poststratification,demsar2006statistical,hsu1996multiple}. We ask what follows once the public data are fixed: for two configurations, what pairwise conclusion is justified, under what target, label source, support condition, uncertainty unit, and decision rule? Our procedure combines these conditions and returns decision labels for agent leaderboard comparisons; to our knowledge, this combination is absent from existing public-log methods.

This distinction between a released score and a pairwise superiority claim leads to three contributions:
\begin{enumerate}
    \item \textbf{An estimand-aware pairwise decision procedure.} The procedure specifies the evaluated configuration, score estimand, label source, support cells, uncertainty calculation, and decision rule needed to interpret a published ordering.
    \item \textbf{Finite-sample labels for leaderboard claims.} The labels separate stable pairwise conclusions from comparisons limited by missing support, verifier disagreement, insufficient precision, or target choice.
    \item \textbf{Controlled and empirical evidence from public agent leaderboards.} We test label behavior under known scenarios, then apply the procedure to three detailed public releases and two coarser public exports to show how conclusions from unadjusted orderings depend on the measurement source, target population, protocol, utility rule, and release detail.
\end{enumerate}

\section{Related Work}

\textbf{Agent benchmarks and evaluation documentation.}
Agent benchmarks define tasks, environments, harnesses, verifiers, and public reporting formats for interactive systems \cite{agentbench,swebench,agentrewardbench,dataagentbench,tau2bench,hal}. Prior work argues that agent evaluations need cost control, reproducibility, careful benchmark design, and structured result documentation \cite{aiagentsthatmatter,efficientagents,evalfactsheets}. Building on this documentation, we examine which pairwise conclusions the reported conditions justify in public logs.

\textbf{Targets, weak labels, and statistical comparison.}
Survey sampling and post-stratification separate a target population from an observed sample \cite{cochran1977sampling,little1993poststratification}, and causal-inference texts make the estimand explicit before discussing identification or estimation \cite{imbensrubin2015}. Calibration weighting, entropy balancing, hierarchical models, and item-response models provide stronger estimators when their assumptions and data support hold \cite{battaglia2009raking,hainmueller2012entropy,gelmanhill2007,lord1980irt,embretson2000irt}. Preference-evaluation methods study human and automatic judgment settings and evaluator biases \cite{mtbench,chatbotarena,fairllmjudge,alpacalength,humaine}. Prediction-powered inference methods show how predictions can be adjusted when paired labels are available \cite{ppi,ppiplus}. These ideas guide our treatment of target choice and paired labels: the appropriate analysis depends on the estimand and the information released with the leaderboard.

\textbf{Leaderboard robustness, utility, and public-data limits.}
Benchmark-target and uncertainty analyses show that conclusions can change under benchmark perturbations, prompt-distribution assumptions, and uncertainty-aware scoring \cite{benchmarktargets,siskabenchrobust,llmuq}. Competition-leaderboard work studies adaptive overfitting and leaderboard operations \cite{blumhardt2015ladder,zhao2025lbops}; classifier-comparison and multiple-comparison methods treat rank differences as inferential claims rather than point estimates alone \cite{demsar2006statistical,hsu1996multiple}. Distributionally robust optimization and sensitivity analysis provide tools for robustness and hidden bias \cite{rahimian2022dro,esfahani2018wasserstein,duchi2021dro,rosenbaum2002observational}. Contextual-bandit studies empirically compare learning algorithms under bandit feedback \cite{bietti2018contextual}. Off-policy evaluation methods make logging overlap and support explicit \cite{narita2019counterfactual,sachdeva2020support}. Multi-criteria decision analysis formalizes utility tradeoffs \cite{keeney1976decisions,belton2002mcda}. Because most public agent leaderboards do not expose the data needed for full causal or off-policy analyses, we focus on the score targets and pairwise conclusions that remain estimable from released logs.

\section{A Pairwise Comparison Procedure}
\label{sec:procedure}

An LLM-agent ordering is interpretable only when the score estimand and the pairwise conclusion refer to the same evaluation target. The procedure asks three questions: what does the score estimate, do the released data cover that target, and is the observed difference large enough for the stated rule? It interprets the published score by reporting the task population, label source, and uncertainty rule under which each pairwise difference can support a conclusion.

As a running example, consider two top SWE-bench submissions with raw success rates 0.792 and 0.788. The raw ordering asks which submission solved more of the 500 released instances. An equal-repository target asks a different question: how the same pair compares if each repository contributes equally. If both submissions have runs in every repository, the target is covered and the pair can be standardized. If the interval for the standardized difference includes zero or the gap does not exceed the 2-point margin, the comparison is underpowered rather than an inferentially supported superiority conclusion. The rest of this section defines this sequence for arbitrary public releases.

\subsection{What Quantity Does the Score Estimate?}

A leaderboard score can estimate observed benchmark success, target-standardized success, weak-verifier success, or utility. The same number can describe a scaffold-specific run on the observed task mix, success after reweighting to a stated task population, a weak-verifier measurement, or a utility rule that trades success against resources such as cost or latency. The analysis records the evaluated unit, the score estimand, and the label or utility source.

The evaluated unit is an agent configuration $a$: base model, scaffold, tools, action space, prompt policy, stopping rule, verifier, and resource budget. A model-only claim requires extra design support, such as multiple scaffolds per model and multiple models per scaffold. Otherwise, the score describes the evaluated configuration.

An estimand is the quantity a reported score is trying to estimate. Let $Y_a$ be binary success for configuration $a$, let $P_0(X)$ be the observed evaluation mixture over pre-run covariates $X$, and let $P^\star(X)$ be a stated target population. The same observed run can support different score definitions. Under $P_0$, the unadjusted average estimates success on the observed mix. Under $P^\star$, a target-standardized score asks how the configuration would score if the task mix changed within supported cells. With paired expert labels, verifier calibration asks what weak-verifier success implies for expert-aligned success. With cost, latency, or action budgets, a utility score asks which configuration is best under a stated resource rule.

These scores answer different questions, so an ordering under one estimand should not be read as an ordering under the others. In the empirical sections, a \emph{capability} claim refers to expert or official success under a target population; a \emph{measurement} claim refers to a weak-verifier label source; and a \emph{utility} claim refers to a stated success-cost rule.

\subsection{Do the Evaluation Data Cover the Target?}

A target can require observations that the evaluation data do not contain. If an equal-domain target assigns weight to a domain where one agent has no runs, the all-agent comparison is not identified without modeling unobserved outcomes. The support check separates the stated target from the part of that target covered for all compared agents.

For agents $\mathcal{A}$, define the common-support set as
\begin{equation}
\mathcal{X}_{\mathrm{common}} =
\{x : n_{a,x} > 0 \textrm{ for every } a \in \mathcal{A}\}.
\end{equation}
If the stated target assigns mass outside this set, the all-agent comparison is not identified without modeling assumptions. The comparison is reported as unsupported, or the target is restricted and the interpretation changes.

The primary estimate is direct standardization:
\begin{equation}
\hat{\psi}^\star(a)=\sum_x \pi^\star(x)\hat{p}_{a,x},
\end{equation}
where $\pi^\star(x)$ are target weights and $\hat{p}_{a,x}$ is the empirical success rate for configuration $a$ in cell $x$. Standardization reports how scores and pairwise differences change under the stated target. It does not attribute those changes to causal effects of cells. When the evaluation data lack support, repeated structure, or item-level records, the analysis is limited to the coarser question the data expose.

\subsection{When Is a Score Difference Large Enough?}

Pairwise score differences are often read as evidence that one configuration outperforms another. We call such a difference a gap. A small gap may disappear under resampling, a practical margin, or a different target mix. Inferential support requires stating both the uncertainty unit and the gap size that is large enough to matter. Let $\hat{\Delta}_{ab}$ be the target-standardized gap for configurations $a$ and $b$, and let $C_{ab}=[L_{ab},U_{ab}]$ be its interval. Under margin $m$, $a$ is a practical winner if and only if $L_{ab}>0$ and $\hat{\Delta}_{ab}\geq m$; $b$ is a practical winner if and only if $U_{ab}<0$ and $\hat{\Delta}_{ab}\leq -m$. Thus, the interval supports the direction, while the margin applies to the point estimate. When item-level outcomes are available, we compare agents on matched tasks. When only grouped results are available, we compare within the reported cells, such as benchmark, repository, dataset, or domain. Bootstrap intervals resample tasks or reported cells according to the dataset's repeated structure; aggregate-only datasets receive grouped stability checks rather than item-level uncertainty claims.

The ordered label names the first condition that blocks a pairwise superiority claim. Identification comes first: a comparison is \emph{unsupported} when the stated target lacks common support. The next question is whether the measurement source changes the conclusion. A comparison is \emph{verifier-sensitive} when expert or official labels and a weak verifier select different winners on the supported target.

For a supported comparison with a fixed label source, the procedure next examines precision. Here, \emph{underpowered} is a compact decision label rather than a post-hoc power calculation: either the interval includes zero, or the target-standardized gap is smaller than the stated practical margin. The label does not imply that increasing the sample size would change the conclusion, particularly when the point estimate remains inside the margin. If precision is sufficient, the comparison is \emph{target-sensitive} when the target yields a supported practical winner opposite the displayed raw ordering or when the uncertainty-supported movement from the released mix to the target exceeds the stated margin. A comparison is \emph{stable} when none of these conditions holds and the target-standardized conclusion excludes zero by the chosen uncertainty rule.

We use a 2 percentage-point reporting margin to avoid treating sub-point and near-tie differences as practical superiority when no release-specific cost or switching rule is available. A release-specific rule takes precedence when one is available. Replicate-level outputs retain all flags because a comparison may satisfy several conditions even when the paper reports one label.

The named-target comparison asks whether the conclusion changes under one specified alternative. We also ask how far the target weights would have to move, within common support, before the pairwise conclusion changes. This calculation gives the minimum target reweighting that would alter the result.

The intuition is ordinary reweighting. If one coding agent leads because the benchmark contains many tasks from repositories where it does well, the calculation asks how much repository weight must shift toward supported repositories where another agent does better.

Let $d_x(a,b)=\hat{p}_{a,x}-\hat{p}_{b,x}$ be the cell-level gap and let $\pi_0$ be the baseline target. If $\hat{\Delta}_{\pi_0}(a,b)>0$, the sign-flip radius is
\begin{equation}
\rho_0(a,b)=
\min_{\pi \in \Delta(\mathcal{X}_{ab})}
\|\pi-\pi_0\|_1
\quad
\textrm{s.t.}
\quad
\sum_x \pi_x d_x(a,b) \leq 0,
\end{equation}
where $\Delta(\mathcal{X}_{ab})$ is the simplex over the pair's common-support cells. We call $\rho_0(a,b)$ the sign-flip radius. If every supported cell favors $a$, then $\rho_0(a,b)$ is infinite. We use the same calculation for a practical margin by replacing $0$ with the stated margin. Smaller values mean that less target mass must move toward cells where $b$ is stronger before $a$'s lead disappears. The radius is an L1 reweighting distance and has no probabilistic or causal interpretation.

For nominal cells, L1 distance has a direct reading as target mass shifted between supported cells. Because L1 distance counts mass removed and mass added, an L1 radius of 0.25 corresponds to shifting 12.5 percentage points of target mass from favorable cells to unfavorable cells.

Algorithm~\ref{alg:support-aware} gives the pairwise procedure. It returns all diagnostic flags for inspection and one ordered label for compact reporting.

\begin{center}
\refstepcounter{algorithm}\label{alg:support-aware}
\begin{minipage}{0.97\columnwidth}
\footnotesize
\hrule
\vspace{0.25em}
\noindent\textbf{Algorithm~\thealgorithm: Public-log pairwise comparison}

\vspace{0.15em}
\noindent\textbf{Input:} configurations $a,b$; outcomes by cell; target weights; label or utility source; margin $m$; support table; resampling rule; optional paired expert and weak labels; optional comparison family and error-control rule.

\vspace{0.15em}
\noindent\textbf{Output:} support status, standardized scores, pairwise difference interval, sign-flip radius when supported, diagnostic flags, and ordered decision label.

\vspace{0.15em}
\begin{enumerate}
\setlength{\itemsep}{0.05em}
\setlength{\parsep}{0pt}
\setlength{\topsep}{0.1em}
    \item State the comparison fields: target, label or utility source, margin $m$, uncertainty unit, resampling rule, and, for family-level claims, the comparison family and error-control rule.
    \item Build common-support cells. If the target puts mass outside common support, set the unsupported flag.
    \item Estimate observed-mix and target-standardized scores on supported cells.
    \item Estimate the pairwise difference and a pointwise interval for a named pair, or a simultaneous or adjusted interval under the prespecified rule for a declared family.
    \item If paired expert and weak labels exist, compare their supported decisions and set the verifier-sensitive flag when they disagree.
    \item Compute the sign-flip or margin-flip radius when cell-level gaps are available.
    \item Set all diagnostic flags: unsupported, verifier-sensitive, underpowered, target-sensitive, and stable.
    \item Return all flags and the first applicable ordered label.
\end{enumerate}

\vspace{0.15em}
\noindent\textbf{Label order:} unsupported, verifier-sensitive, underpowered, target-sensitive, stable. All diagnostic flags are retained, but the compact label reports the earliest blocking issue. When bootstrap uncertainty is available, a comparison is target-sensitive by gap size only if the lower bound on the absolute change in the pairwise gap exceeds $m$.
\vspace{0.2em}
\hrule
\end{minipage}
\end{center}

These fields define the score's target estimand and the evidence available for pairwise conclusions. A prespecified named-pair claim can use a pointwise interval. A tier-wide or winner claim must also declare its comparison family and error-control target, so the correction matches the claim and dependence structure \cite{hsu1996multiple}.

\section{Controlled Behavior Checks}
\label{sec:controlled-checks}

Because public leaderboard logs do not reveal a ground-truth ordering, we first examine the procedure in controlled reference cases. Simulations assess label recovery and whether standardization and weak-verifier calibration reduce error when their assumptions hold. An AgentRewardBench permutation analysis then provides a task-conditioned null reference for the observed balanced gaps. These checks establish how the labels behave before we turn to the public releases.

\subsection{Decision-Label Recovery}

The label-recovery simulation uses two configurations and four evaluation cells. Each scenario fixes the true cell success probabilities, the observed cell mix, the target mix, support pattern, verifier behavior, and practical margin. We then draw Bernoulli outcomes and apply the procedure without using the true label. Table~\ref{tab:controlled-labels} reports 300 replicates per scenario with 300 bootstrap draws per replicate. The modal label matches the known state in every scenario, although finite sampling produces occasional classification errors.

\begin{table}[t]
\centering
\footnotesize
\caption{Controlled behavior of decision labels. Intervals are Wilson 95 percent intervals over replicates.}
\label{tab:controlled-labels}
\setlength{\tabcolsep}{2pt}
\begin{tabular}{@{}L{0.22\columnwidth}L{0.39\columnwidth}L{0.31\columnwidth}@{}}
\toprule
Scenario & Known state & Recovery \\
\midrule
Stable winner & One configuration leads in every supported cell. & 99.7\% [98.1, 99.9] \\
Target-sensitive & Observed mix favors one configuration; equal-cell target favors the other. & 98.3\% [96.2, 99.3] \\
Unsupported & Target assigns mass to a cell missing for one configuration. & 100.0\% [98.7, 100.0] \\
Verifier-sensitive & Expert labels favor one configuration; weak labels favor the other. & 100.0\% [98.7, 100.0] \\
Underpowered & True target gap is below the 2-point margin. & 95.7\% [92.7, 97.5] \\
\bottomrule
\end{tabular}
\end{table}

The same simulation tests the target-sensitivity rule. A simpler rule that labels target sensitivity from point-estimate movement alone recovered the stable scenario in only 84.3 percent of replicates, overcalling target sensitivity in 47 of 300 runs. The uncertainty-aware rule in Section~\ref{sec:procedure} recovered the stable scenario in 99.7 percent of replicates, so the empirical analysis uses uncertainty-supported movement when item-level or cell-level uncertainty is available.

\subsection{Known-Target Estimation Check}

The known-target simulation asks whether the estimators reduce error when the target value is known. We simulate four agent configurations evaluated across assistant, web, visual, and work conditions, with benchmark counts drawn from an AgentRewardBench-inspired mixture $(0.45,0.30,0.05,0.20)$; the target gives equal weight to the four benchmarks. Direct standardization reduces mean absolute error from 0.0184 for the raw observed expert score to 0.0128. Regression standardization gives 0.0139, and cross-fitted weak-verifier calibration gives 0.0144. By comparison, uncalibrated weak-verifier estimates have mean absolute error 0.0401 after direct standardization and 0.0804 for the raw weak-verifier score. In a separate missing-support scenario, the full equal-benchmark target is unsupported in every replicate for the agent missing visual observations, while the common-support target remains estimable.

\subsection{Permutation Check on a Public Release}

On AgentRewardBench, we permute agent labels within task clusters. This preserves task difficulty and benchmark composition while testing a null in which agent identity is exchangeable after conditioning on task. Across 1,000 permutations, the observed maximum absolute balanced pairwise gap is 0.1489, while the permutation 95th percentile is 0.0727. The plus-one corrected empirical $p$ value for the maximum gap is below 0.001. The count of practical gaps at least 2 percentage points also exceeds the permutation reference, with plus-one corrected empirical $p$ value 0.010.

We also perturb the common-support target weights around equal benchmark weights. Claude remains the top agent in all 1,000 perturbation draws, but the 10th percentile of the top gap is 0.0165. The top position is stable near the equal-benchmark target, while some of its pairwise margins remain small.

\section{Empirical Evidence}
\label{sec:empirical}

\subsection{What the Public Releases Contain}

The public records differ in resolution, so we match each comparison to the finest analysis its release supports. We apply the same ordered label rule across releases while adapting the uncertainty calculation to each export's repeated structure. SWE-bench provides repository covariates and instance-level outcomes for target reweighting and bootstrap uncertainty \cite{swebench,swebenchexperiments}. AgentRewardBench pairs expert and weak labels at the trajectory level, which permits common-support and verifier analyses \cite{agentrewardbench,agentrewardbenchdata}. tau2-bench adds repeated attempts, run modes, task metadata, cost, and duration, supporting separate protocol and utility comparisons \cite{tau2bench,tau2benchdata}. DataAgentBench and Open Agent provide coarser records, so their analyses are limited to the score reweighting and grouped comparisons those exports support \cite{dataagentbench,dataagentbenchdata,openagentleaderboarddata}. Variables and dependence structures absent from an export remain outside the corresponding analysis.

\subsection{SWE-bench: Most Top-10 Pairwise Orderings Are Unresolved}

SWE-bench Verified lets us ask whether a close ordering survives a change in the target population. The public release covers 134 submissions, 500 instances, and 12 repositories \cite{swebenchexperiments}. The raw score is the fraction of instances resolved, whereas the target-standardized score gives each repository equal weight. Equal-repository scoring is a plausible alternative target for code-agent leaderboards because repository composition is visible and uneven. We ask whether a close unadjusted ordering still has inferential support under that stated target.

Repository standardization changes practical conclusions. Two submissions tie at raw success 0.792. The Live-SWE-agent Claude 4.5 Opus medium submission has repository-balanced success 0.768, with bootstrap interval $[0.690,0.833]$. The top repository-balanced submission, TRAE with Doubao-Seed-Code, has raw success 0.788 and balanced success 0.780, with interval $[0.694,0.846]$. Across all submissions, the maximum absolute shift is 12.0 points.

\begin{figure}[t]
\centering
\includegraphics[width=\columnwidth]{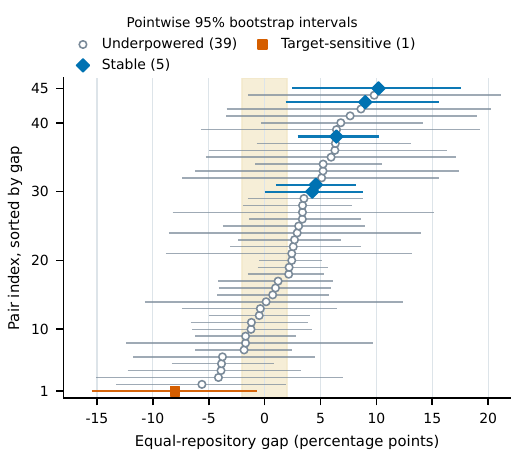}
\caption{Equal-repository gaps for the 45 pairs formed conditional on the observed raw top 10 SWE-bench submissions, sorted by point estimate. Points show gaps, horizontal lines show descriptive pointwise 95 percent bootstrap intervals, and the shaded region marks the 2-point practical margin, all in percentage points. Marker shape and color denote ordered decision labels.}
\label{fig:decision-sensitivity}
\end{figure}

Figure~\ref{fig:decision-sensitivity} reports descriptive pointwise uncertainty-aware labels for the 45 pairs formed conditional on the observed raw top 10: 5 are stable under the 2-point practical-margin rule, 1 is target-sensitive, and 39 are underpowered. Point-estimate target movement occurs often, but bootstrap uncertainty usually leaves the pairwise superiority claim unresolved. The labels therefore describe how well the released records distinguish submissions within the observed raw top 10, rather than overall system quality.

Equal-repository reweighting changes many estimated gaps. Among the 45 pairs conditional on the observed raw top 10, 38 move by at least 1 point, 31 by at least 2 points, and 7 by at least 5 points under equal-repository scoring. When these 45 pairs are treated as one comparison family, the max-deviation simultaneous bootstrap interval does not exclude zero for any pair. Across these target and margin choices, the public evidence rarely resolves close unadjusted orderings within the observed raw top 10.

\subsection{AgentRewardBench: Coverage Limits the Target and Label Source Changes the Winner}

AgentRewardBench tests two questions: whether the same benchmark target is covered for all agents, and whether weak verifiers estimate the same quantity as expert labels. We analyze 1,302 deduplicated trajectories across four agents and four analysis groups: AssistantBench, VisualWebArena, WebArena, and a WorkArena group that pools WorkArena with WorkArena++ \cite{agentrewardbench,agentrewardbenchdata}. Expert annotations define the capability label in this release; the functional evaluator and GPT-4o-mini judge are proxy labels.

Common support changes the target before any score adjustment. A full all-agent target over all four analysis groups is unsupported because Llama has no VisualWebArena observations. The common-support target includes AssistantBench, WebArena, and WorkArena. Under this target, ranks do not change, but every score decreases. Claude 3.7 Sonnet moves from 0.339 raw success to 0.300 benchmark-balanced success, with balanced 95 percent interval $[0.250,0.356]$; GPT-4o moves from 0.328 to 0.276, $[0.220,0.335]$; Qwen2.5-VL moves from 0.236 to 0.187, $[0.151,0.230]$; and Llama 3.3 moves from 0.181 to 0.154, $[0.109,0.208]$. The largest absolute shift is 5.1 points.

\begin{figure}[t]
\centering
\includegraphics[width=\columnwidth]{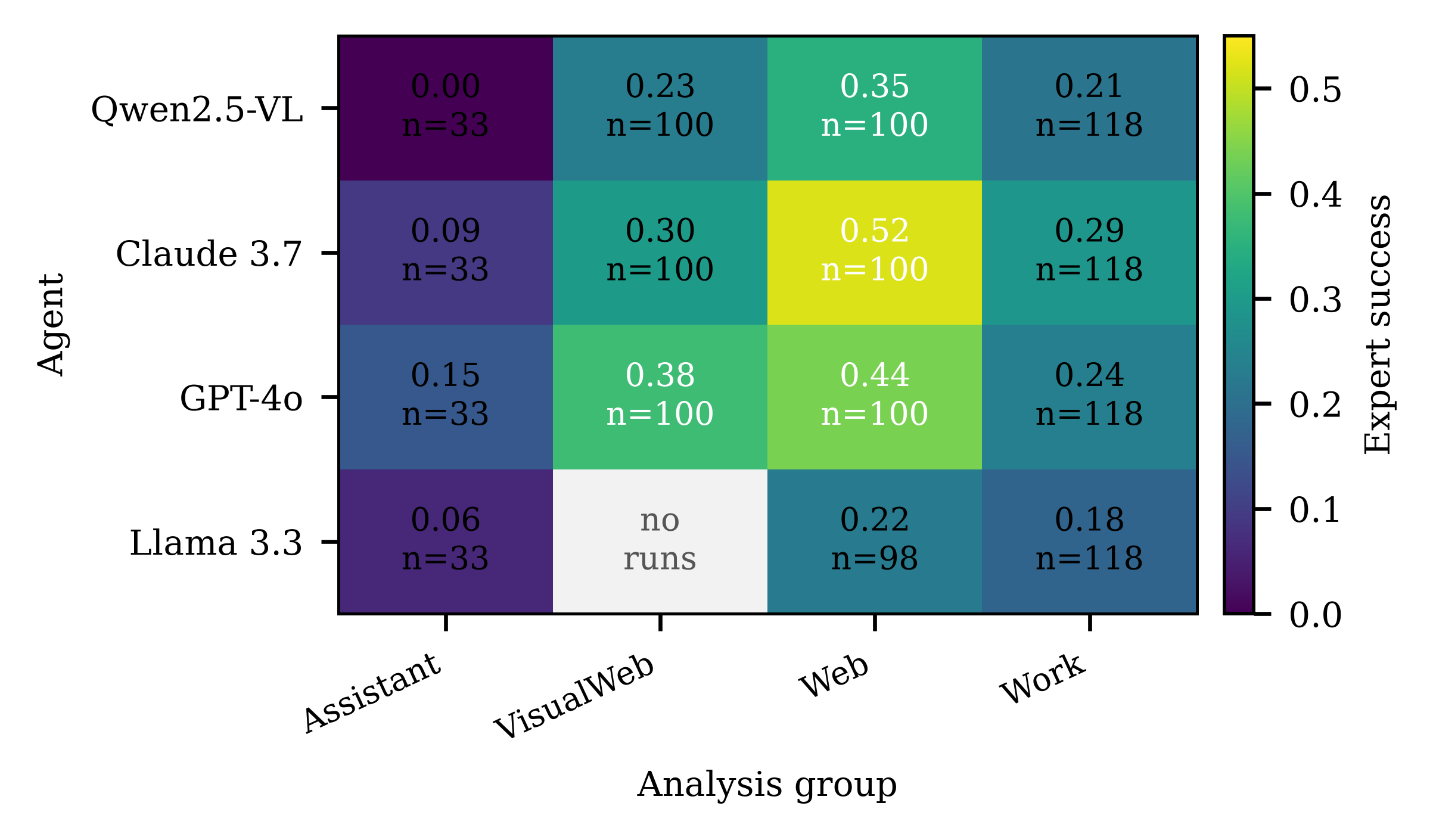}
\caption{Expert success rates by agent and analysis group in AgentRewardBench. Cells report the success rate and trajectory count; ``no runs'' marks missing support.}
\label{fig:arb-heatmap}
\end{figure}

Figure~\ref{fig:arb-heatmap} reports benchmark-specific observed rates: Claude has the highest observed rate on WebArena, GPT-4o has the highest observed rates on AssistantBench and VisualWebArena, and Llama cannot be compared on the visual target without additional assumptions. Under the uncertainty-aware rule, four of six common-support pairwise comparisons are stable and two are underpowered. Benchmark mixture can change score levels without reversing the supported order.

Verifier calibration answers a different question. The functional evaluator underestimates expert success by 8.6 points and has a false negative rate (FNR) of 0.423, while GPT-4o-mini overestimates expert success by 7.9 points and has a false positive rate (FPR) of 0.180. Expert labels select Claude as the top agent, while GPT-4o-mini labels select GPT-4o. Each weak verifier also reverses one of the six pairwise signs relative to expert labels. A leaderboard using either proxy label source measures that proxy, not expert-aligned success.

Paired weak-verifier and expert labels make direct calibration possible in AgentRewardBench. We fit fivefold calibration models by benchmark-task trajectory group, using weak-verifier outputs and benchmark source to predict held-out expert success. The functional and GPT-4o-mini calibrated means are 0.279 and 0.278 against an expert rate of 0.278; their Brier scores are 0.115 and 0.093, and their AUCs are 0.870 and 0.918. The split holds out task groups but allows the same agents to appear in training and test folds, which motivates the leave-one-agent-out analysis below.

To examine whether calibration depends on observing the same agents during training, we repeat the analysis in an exploratory leave-one-agent-out design over the four observed configurations and three common-support benchmarks. Calibration lowers equal-benchmark Brier loss relative to raw weak-success labels in all eight verifier-agent folds. Averaged equally across agents, the calibrated-minus-raw differences are $-0.037$, with descriptive 95 percent task-group bootstrap interval $[-0.053,-0.022]$, for the functional verifier and $-0.052$, with interval $[-0.069,-0.035]$, for GPT-4o-mini. Because tasks recur across agents, these results describe held-out prediction for the four observed configurations on the observed task and benchmark distributions; transport to new models, scaffolds, task or benchmark distributions, or failure distributions would require separate evaluation. Expert-aligned calibration also requires paired expert and proxy labels; releases without such labels permit only comparisons between label sources.

\subsection{tau2-bench: Protocol and Utility Choices Change Selections}

tau2-bench evaluates conversational agents in simulated service workflows \cite{tau2bench}. We analyze 44 non-example public submissions and summarize 10,832 trial rows from 26 publicly released result files \cite{tau2benchdata}. The public submissions are uneven across domains: retail appears in 34 submissions, airline in 33, telecom in 33, and banking knowledge in 18. These domain counts determine which targets are covered.

For task-level comparisons, we restrict the main analysis to four system configurations with common support over airline, retail, and telecom under their published primary modes. We call each separately reported tau2-bench run setting a mode and use the published primary mode for the main comparison. Equal-domain reweighting is smaller than in SWE-bench or DataAgentBench: the largest shift is 2.2 points, yet all six pairwise comparisons remain underpowered under the paired domain-task bootstrap.

Repeated trials, run modes, task covariates, and utility each change a different part of the comparison, so we analyze them separately. Averaging pass@4 over domains raises common-support main-mode scores by 17.7 to 20.4 points relative to pass@1, and the largest pass@4 lift across all domain settings is 24.6 points. Across four telecom modes, o4-mini ranges from 0.421 to 0.989 pass@1, a 56.8 point range.

The logistic adjustment leaves the top system unchanged and moves equal-domain scores by at most 1.58 points relative to direct standardization. Sensitivity to the task distribution depends more strongly on how distance between tasks is defined. With a Gower distance over mixed-type task metadata, the Claude 3.7 Sonnet top decision has a Wasserstein radius of 0.00426 against o4-mini. The radius therefore describes sensitivity under this particular task representation.

Cost can change the selected system under a stated utility rule. For the four common-support primary-mode configurations, with utility equal to success minus $\lambda$ times agent-side cost and duration weight $\mu=0$, the first point-estimate cost-only switch is from Claude 3.7 Sonnet to o4-mini at $\lambda=0.0514$ success-rate units per USD; agent-side cost is measured in USD per trajectory. The utility winner therefore depends on the stated resource rule as well as capability.

\subsection{What Coarser Releases Still Support}

DataAgentBench provides a query-level example. The top public configuration remains PromptQL Claude Opus 4.6 under equal-dataset scoring, but scores shift by up to 9.2 points and 10 of 28 public-configuration pairs remain underpowered \cite{dataagentbench,dataagentbenchdata}. An unchanged top position can therefore coexist with shifted score levels and uncertain lower-pair decisions.

Open Agent provides only aggregate rows. Its top configuration remains unchanged under benchmark standardization, with success 0.784 under the released mix and 0.787 under equal benchmark weights; the largest shift is 2.7 points, and all 66 grouped comparisons among the top 12 configurations are stable \cite{openagentleaderboarddata}. The export does not support item-level uncertainty or verifier calibration, and utility remains a separate question.

The releases reveal distinct limits on raw orderings: unresolved SWE-bench differences, label disagreement in AgentRewardBench, protocol and utility dependence in tau2-bench, and granularity limits in coarser exports. An unadjusted score remains valid for the released mix, but broader claims require corresponding support, labels, repeated trials, or item-level records.

\section{Discussion and Limitations}

Leaderboard authors can make these distinctions visible by reporting what configuration was evaluated, which population and labels define the score, and how support, uncertainty, and practical significance govern close comparisons. When cost, latency, repeats, or scaffolds matter, those quantities should be reported as separate decision dimensions rather than silently folded into one ordering.

The public releases support pairwise conclusions only for the variables and outcomes recorded in them. They do not identify causal effects of repositories, benchmarks, scaffolds, modes, or verifiers. AgentRewardBench permits calibration because it contains paired expert and weak labels, whereas most releases permit only comparisons between label sources.

Sensitivity and multiplicity choices determine the scope of any reported winner. Sign-flip radii describe movement over common-support cells under a stated target distance, while Wasserstein variants also depend on how distance between tasks is defined. The appropriate multiplicity rule follows the claim: a named pair can use a pointwise interval, whereas a family-level winner requires a declared comparison family and error-control rule. Interpretations based on sensitivity or family-level comparisons should therefore state the corresponding target distance or comparison family.

\section{Conclusion}

Across public LLM-agent releases, close ranks often do not support a pairwise winner once the target, common support, and uncertainty are specified. Label sources, repeated-attempt protocols, and utility rules can also change the score or selected system. A leaderboard score summarizes one released evaluation; pairwise superiority additionally requires its estimand and decision rule.

\section*{Acknowledgment}

OpenAI ChatGPT assisted with prose refinement. The author verified all results, citations, and claims and takes responsibility for the manuscript.

\bibliographystyle{IEEEtran}
\bibliography{references}

\end{document}